# Vision-based Unscented FastSLAM for Mobile Robot

Chunxin Qiu, Xiaorui Zhu, *Member, IEEE*, Xiaobing Zhao

*Abstract*—This paper presents a vision-based Unscented FastSLAM (UFastSLAM) algorithm combing the Rao-Blackwellized particle filter and Unscented Kalman filte(UKF). The landmarks are detected by a binocular vision to integrate localization and mapping. Since such binocular vision system generally inherits larger measurement errors, it is suitable to adopt Unscented FastSLAM to improve the performance of localization and mapping. Unscented FastSLAM takes advantage of UKF instead of the linear approximations of the nonlinear function where the effective number of particles is used as the criteria to reduce the particle degeneration. Simulations and experiments are carried out to demonstrate that the Unscented FastSLAM algorithm can achieve much better performance in the vision-based system than FastSLAM2.0 algorithm on the accuracy and robustness.

## I. INTRODUCTION

In the past decade, simultaneous localization and mapping (SLAM) has been the focus of the robot navigation. SLAM problem arise when the robot does not have access to a map of the environment, nor does it know its own pose. Generally, SLAM is a complex problem because the robot requires a good pose estimate while a consistent map is needed simultaneously to localize the robot [1-3].

Early work on SLAM has been done by Smith, Self and Cheeseman [4]. These authors proposed the use of an Extended Kalman Filter (EKF) to consider uncertainties in robot pose and the map. The EKF-SLAM has been used successful in different environments, involving robotic vehicles in the air [5], on the ground [6-7] and underwater [8]. Recent research works have focused on applying to larger-scale environments with more than a few hundred landmarks [6, 7, 9] and on the algorithms for handling data association problems [10]. However, the EKF-SLAM algorithm suffers from its enormous update complexity, and the linearization of the robot and sensor models. These assumptions can be problematic when dealing with large uncertainties and strong non-linearities. To solute the linearization problem, Martinez-Cantin et al. [11] used the unscented Kalman filter (UKF) in SLAM problems. This approach was used to avoid the analytical linearization based on Taylor series expansion of the nonlinear models, and improved the consistency over the EKF-based approach.

Murphy [12] introduced the Rao-Blackwellized particle filter (RBPF) as an efficient solution to the SLAM problem. In the RBPF, each particle represents a potential trajectory of the robot and a map of the environment. One of the major challenges for the RBPF is to reduce the number of particles while maintaining the estimation accuracy. Montemerlo et al. [13-15] subsequently developed the RBPF framework into a robot localization problem and a collection of landmark estimation problems that are conditioned on the robot pose estimate. RBPF-SLAM is also called FastSLAM due to its computational advantages over the EKF SLAM. However, FastSLAM still needs to deal with the nonlinear function by deriving the Jacobian matrices that could result in the filter inconsistency.

The Unscented FastSLAM algorithm was proposed to overcome the drawbacks of FastSLAM where the scaled unscented transformation (SUT) was applied to replace the linearization in the FastSLAM framework [16]. Kim et al. [17] used the laser scanner as the range sensor to demonstrate the performance of UFastSLAM, and also used the sonar sensor to validate the robustness of UFastSLAM. In the last several years, vision sensors have been readily used on navigation of mobile robot due to their low-cost property. However, more accurate feature extraction from the images captured by vision sensors is very time-consuming, which blocks most online applications. On the other hand, rough image process with less computational cost could result in failure of SLAM because of large measurement errors. Therefore, this paper investigates how to improve accuracy and robustness of localization and mapping using UFastSLAM.

This paper is organized as follows. Robot model is presented in Section II. Unscented FastSLAM algorithm is presented in Section III. Simulations and experiments are presented in Section IV for verification. Finally, conclusions are described in Section VI.

## II. ROBOT MODEL

*A. Motion model*

Two coordinate systems are used in the paper. They are the world coordinate system ($o_w, x_w, y_w$) and the robot coordinate system ($o_r, x_r, y_r$). The pose of a mobile robot operating in a plane is illustrated as $(x, y, \theta)$. Let $u_t = (v \quad w)^T$ denote the control at time t. If the control input are kept at a fixed value during the time interval $(t-1, t]$, the robot moves with the

This research is supported by the NSF China Grant No. 60905052.
Xiaorui Zhu is the corresponding author with Harbin Institute of Technology Shenzhen Graduate School, Shenzhen, Guangdong 518055, China (e-mail: xiaoruizhu@hitsz.edu.cn).

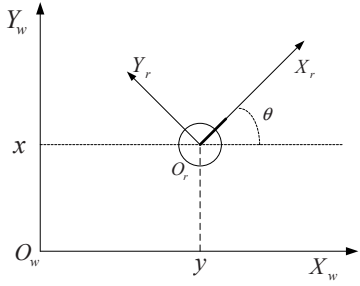

Fig.1 Robot motion model

radius $r = |v/w|$. Let $x_{t-1} = (x \quad y \quad \theta)^T$ be the initial pose of the robot, in Fig.1. The center of the circle is at:

$$x_c = x - \frac{v}{w}\sin\theta \qquad (1)$$

$$y_c = y + \frac{v}{w}\cos\theta \qquad (2)$$

Due to the measurement errors, the actual velocities of the robot are given by [18]:

$$\begin{pmatrix}\hat{v}\\\hat{w}\end{pmatrix} = \begin{pmatrix}v\\w\end{pmatrix} + \begin{pmatrix}\xi_{\alpha_1|v|+\alpha_2|w|}\\\xi_{\alpha_3|v|+\alpha_4|w|}\end{pmatrix} \qquad (3)$$

where $\xi$ is a zero-mean error variable. The parameters $\alpha_1 \sim \alpha_4$ are the control error of the robot. The less accurate a robot, the larger these parameters. Thus, the resulting motion model is as follows:

$$\begin{bmatrix}x'\\y'\\\theta'\end{bmatrix} = \begin{bmatrix}x\\y\\\theta\end{bmatrix} + \begin{bmatrix}-\frac{\hat{v}}{\hat{w}}\sin\theta + \frac{\hat{v}}{\hat{w}}\sin(\theta+\hat{w}\Delta t)\\\frac{\hat{v}}{\hat{w}}\cos\theta - \frac{\hat{v}}{\hat{w}}\cos(\theta+\hat{w}\Delta t)\\\hat{w}\Delta t\end{bmatrix} \qquad (4)$$

where $x_t = (x' \quad y' \quad \theta')^T$ is the actual pose after executing the motion command $u_t = (v \quad w)^T$ at $x_{t-1} = (x \quad y \quad \theta)^T$, and $\Delta t$ is the time interval.

*B. Feature-based Measurement model*

The feature-based map is used in this paper. The features are measured by a binocular vision system including the range $r$ and the bearing $\varphi$ of the landmark relative to the robot coordinate frame of the robot. In this paper, each landmark has different color such that the color can be treated as the identity of each landmark. Then the measurement model can be described as,

$$\begin{bmatrix}r\\\varphi\end{bmatrix} = \begin{bmatrix}\sqrt{(m_{i,x}-x)^2+(m_{i,y}-y)^2}\\a\tan 2(m_{i,x}-x, m_{i,y}-y) - \theta\end{bmatrix} + \begin{bmatrix}\varepsilon_{\sigma_r^2}\\\varepsilon_{\sigma_\varphi^2}\end{bmatrix} \qquad (5)$$

Here $(m_{i,x}, m_{i,y})$ is the coordinate of $i^{th}$ feature in the global coordinate frame of the map. The robot pose is given by $x_t = (x \quad y \quad \theta)^T$ at time $t$. $\varepsilon_{\sigma_r}$, $\varepsilon_{\sigma_\phi}$ are zero-mean Gaussian error variables with standard deviations $\sigma_r$, $\sigma_\phi$, respectively.

### III. UNSCENTED FASTSLAM ALGORITHM

Unscented FastSLAM integrates the scaled unscented transformation (SUT) into the SLAM framework. There are three steps including predicting the robot pose, estimating the feature position, and calculating the importance weights and resampling, Table 1.

*A. Robot pose estimation*

At first, the state vector is augmented with a control input and the observation,

$$x_{t-1}^{a[k]} = [x_{t-1}^{[k]} \quad 0 \quad 0]^T \qquad (6)$$

$$P_{t-1}^{a[k]} = \begin{bmatrix}P_{t-1}^{[k]} & 0 & 0\\0 & R_t & 0\\0 & 0 & Q_t\end{bmatrix} \qquad (7)$$

where $x_{t-1}^{a[k]}$ is the augmented vector that includes the robot pose, the control input and the observation. $Q_t$ and $R_t$ are the control noise covariance and the measurement noise covariance, respectively. The augmented covariance matrix $P_{t-1}^{a[k]}$ has 7x7 dimensions.

Unscented FastSLAM deterministically extracts a set of $2N+1$ sigma points from the mean point. $N=7$ is the augmented state vector and the sigma points can be calculated as below,

$$\chi_{t-1}^{a[0][k]} = x_{t-1}^{a[k]} \qquad (8)$$

$$\chi_{t-1}^{a[i][k]} = x_{t-1}^{a[k]} + (\sqrt{(N+\lambda)P_{t-1}^{a[k]}})_i \quad (i=1,...,N) \qquad (9)$$

$$\chi_{t-1}^{a[i][k]} = x_{t-1}^{a[k]} + (\sqrt{(N+\lambda)P_{t-1}^{a[k]}})_{i-N} \quad (i=L+1,...,2N) \quad (10)$$

where $\lambda = \alpha^2(n+k) - n$, $\alpha$ and $k$ are constant. Each sigma point contains the robot pose, the control input, the observation. The set of the sigma points is transformed by the motion model as follow,

$$\bar{\chi}_t^{[i][k]} = f(u_t, \chi_{t-1}^{[i][k]}) \qquad (11)$$

So the predicted values of the mean and covariance of the robot pose are calculated as,

$$x_{t|t-1}^{[k]} = \sum_{i=0}^{2N} w_m^{[i]} \bar{\chi}_t^{[i][k]} \qquad (12)$$

$$P_{t|t-1}^{[k]} = \sum_{i=0}^{2N} w_c^{[i]} (\bar{\chi}_t^{[i][k]} - x_{t|t-1}^{[k]})(\bar{\chi}_t^{[i][k]} - x_{t|t-1}^{[k]})^T \qquad (13)$$

where $w_m^{[i]}$, $w_c^{[i]}$ are used to compute the mean and covariance. $w_m^{[i]}$ and $w_c^{[i]}$ are calculated by

$$w_m^{[0]} = \frac{\lambda}{n-\lambda} \tag{14}$$

$$w_c^{[0]} = \frac{\lambda}{n+\lambda} + (1-\alpha^2+\beta) \tag{15}$$

$$w_c^{[i]} = w_m^{[i]} = \frac{1}{2(n+\lambda)} \text{ for } i=1,...,2n \tag{16}$$

Here $\beta$ is used to incorporate the knowledge of the higher order moments of the posterior distribution. For a Gaussian prior, the optimal choice is $\beta$ =2 [19].

To obtain more accurate pose estimation, the features are used to update the predicted the mean and covariance of the robot pose:

$$\overline{N}_t^{[i][k]} = h(\overline{\chi}_t^{[i][k]}, \mu_{k,t-1}^k) \tag{17}$$

$$\hat{n}_t^{[k]} = \sum_{i=0}^{2N} w_m^{[i]} \overline{N}_t^{[i][k]} \tag{18}$$

$$S_t^{[k]} = \sum_{i=0}^{2N} w_c^{[i]} (\overline{N}_t^{[i][k]} - \hat{n}_t^{[k]})(\overline{N}_t^{[i][k]} - \hat{n}_t^{[k]})^T \tag{19}$$

Here $\overline{N}_t^{[i][k]}$ are the sigma points, $\hat{n}_t^{[k]}$ is the predicted measurement and $S_t^{[k]}$ is the updated covariance. The cross-covariance and the Kalman gain can be obtained as:

$$\sum\nolimits_t^{x,n[k]} = \sum\nolimits_{i=0}^{2N} w_c^{[i]} (\overline{\chi}_t^{[i][k]} - x_{t|t-1}^k)(\overline{N}_t^{[i][k]} - \hat{n}_t^{[k]})^T \tag{20}$$

$$K_t^{[k]} = \sum\nolimits_t^{x,n[k]} (S_t^{[k]})^{-1} \tag{21}$$

The estimated mean and covariance of the robot state at time $t$ are calculated by

$$x_t^{[k]} = x_{t|t-1}^k + K_t^{[k]}(z_t - \hat{n}_t^{[k]}) \tag{22}$$

$$P_t^{[k]} = P_{t|t-1}^k - K_t^{[K]} S_t^{[k]} (K_t^{[K]})^T \tag{23}$$

Using the updated mean and covariance, the sigma point is updated as,

$$\chi_t^{a[i][k]} = [x_t^{a[k]}, x_t^{a[k]} \pm \sqrt{(N+\lambda)P_t^{a[k]}}] \tag{24}$$

### B. Feature estimation

The estimation of each landmark location follows the same procedure as that of the robot pose. The set of sigma points should be constructed first as,

$$\chi^{a[0][k]} = \mu_{n_t,t-1}^{[k]} \tag{25}$$

$$\chi^{a[i][k]} = \mu_{n_t,t-1}^{[k]} + (\sqrt{(n+\lambda)\sum\nolimits_{n_t,t-1}^m})_i \ (i=1,...,n) \tag{26}$$

Table 1 Unscented FastSLAM Algorithm

| | |
|---|---|
| 1 | initialization parameters |
| 2 | for k=1 to M |
| 3 |    retrieve the robot's pose $x_{t-1}^{[k]}$ from $\chi_{t-1}$ |
| 4 |    predict mean and covariance of robot's pose |
| 5 |    observation data association |
| 6 |    for observed features |
| 7 |       update mean and covariance of the robot's pose |
| 8 |       update the sigma points |
| 9 |       calculate the importance weight |
| 10 |    endfor |
| 11 |    if new features |
| 12 |       initialization this feature |
| 13 |    else |
| 14 |       update this feature |
| 15 |    endif |
| 16 |    for unobserved features |
| 17 |       $\mu_{n_t,t}^{[k]} = \mu_{n_t,t-1}^{[k]}, \sum_{n_t,t}^{[k]} = \sum_{n_t,t-1}^{[k]}$ |
| 18 |    endfor |
| 19 |    add the $\langle x_t^{[k]}, N_t^{[k]}, \mu_{N_t^{[k]},t}^k, \sum_{N_t^{[k]},t}^k \rangle$ to the $\hat{\chi}_t$ |
| 20 | endfor |
| 21 | for k=1 to M |
| 22 |    normalize weight and calculate the $w_{neff}$ |
| 23 |    if $w_{neff} < w_\lambda$ |
| 24 |       resample |
| 25 |    else |
| 26 |       maintain the original particle weight |
| 27 | endfor |
| 28 | obtain the $\chi_t$ |

$$\chi^{a[i][k]} = \mu_{n_t,t-1}^{[k]} - (\sqrt{(n+\lambda)\sum\nolimits_{n_t,t-1}^m})_{i-n} \ (i=1,...,2n) \tag{27}$$

where $\mu_{n_t,t-1}^{[k]}$ and $\sum_{n_t,t-1}^{[m]}$ are the mean and covariance of $n^{th}$ feature at time $t$-$1$. The predicted measurement $\hat{z}_t^{[k]}$ and the covariance $\overline{S}_t^{[k]}$ can be calculated as,

$$\overline{N}_t^{[i][k]} = h(\chi_t^{[i][k]}, x_t^k) \ (i=0,...\ 2n) \tag{28}$$

$$\hat{z}_t^{[k]} = \sum_{i=0}^{2N} w_m^{[i]} \overline{Z}_t^{[i][k]} \tag{29}$$

$$\overline{S}_t^{[k]} = \sum_{i=0}^{2N} w_c^{[i]} (\overline{Z}_t^{[i][k]} - \hat{z}_t^{[k]})(\overline{Z}_t^{[i][k]} - \hat{n}_t^{[k]})^T \tag{30}$$

Table 2. Estimation errors with different measurement errors.

| No. | $\sigma_r$ (m) | $\alpha_\varphi$ (º) | Max-pose error(m) (UFastSLAM) | Max- pose error(m) (FastSLAM2.0) |
|---|---|---|---|---|
| 1 | 0.1 | 1 | 0.86 | 0.55 |
| 2 | 0.3 | 3 | 1.50 | 1.09 |
| 3 | 0.6 | 6 | 2.40 | 1.35 |

where the sigma points $\overline{N}_t^{[i][k]}$ are obtained by the nonlinear transformation. Then the mean and covariance of the landmark location can be updated using Kalman gain as,

$$\Sigma_t^{[k]} = \sum_{i=0}^{2N} w_c^{[i]}(\chi^{[i][k]} - \mu_{n_t,t-1}^{[k]})(\overline{Z}_t^{[i][k]} - \hat{z}_t^{[k]})^T \quad (31)$$

$$\overline{K}_t^{[k]} = \overline{\Sigma}_t^{[k]}(\overline{S}_t^{[k]})^{-1} \quad (32)$$

$$\mu_{n_t,t}^{[k]} = x_{n_t,t-1}^k + \overline{K}_t^{[k]}(z_t - \hat{z}_t^{[k]}) \quad (33)$$

$$\Sigma_{n_t,t}^{[k]} = \Sigma_{n_t,t-1}^{[k]} - \overline{K}_t^{[K]}\overline{S}_t^{[k]}(\overline{K}_t^{[K]})^T \quad (34)$$

*C. Calculating the Importance Weights and Resampling*

The importance weight is obtained as follows:

$$w_t^{[k]} = \left|2\pi L_t^{[t]}\right|^{-\frac{1}{2}} \exp\left\{-\frac{1}{2}(z_t - \overline{z}_t)L_t^{[t-1]}(z_t - \overline{z}_t)\right\} \quad (35)$$

where $L_t^{[t]} = G_x R_t G_x^T + G_m \sum_{c_t,t-1}^{[k]} G_m^T + R_t$ is the covariance. The effective number of particles $N_{eff}$ needs to be calculated to evaluate how well the current particle set represents the true posterior. Referred to Doucet et al. [16], this effective number is computed as,

$$N_{eff} = \frac{1}{\sum_{i=1}^{N}(w_t^{(i)})^2} \quad (36)$$

IV. SIMULATIONS AND EXPERIMENTS

*A. Simulation Results*

To compare the performance between Unscented FastSLAM and FastSLAM2.0, two algorithms were simulated in Matlab®. The experiment area is approximately $100m \times 100m$, and the entire distance of the robot motion is $156m$. The translational velocity variation is chosen as $\sigma_v = 0.02\,m/s$ while the rotational velocity variation is $\sigma_w = 0.2^o\,m/s$. The $\sigma_r$, $\alpha_\varphi$ are the measurement noises of the range and the bearing respectively.

Simulation results are shown in Fig. 2 and Fig. 3 for $\sigma_r = 0.3m$, $\alpha_\varphi = 3^o$. According to Fig. 3, the estimation errors of UFastSLAM increase more slowly than that of FastSLAM2.0.

Three different settings are simulated to test how the measurement uncertainty affects the performance of Unscented FastSLAM comparing with FastSLAM2.0. The average of the pose error is calculated over ten independent runs for each algorithm. As the measurement error was increased, the estimation error of Unscented FastSLAM is smaller than that of FastSLAM2.0, Table 2.

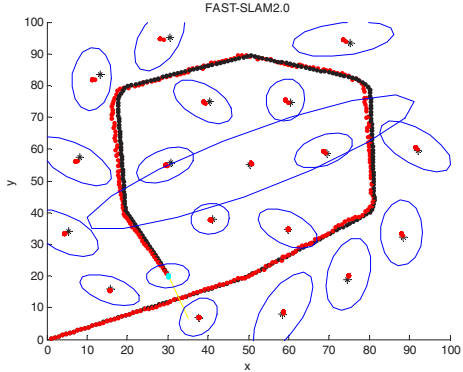

(a) FastSLAM2.0

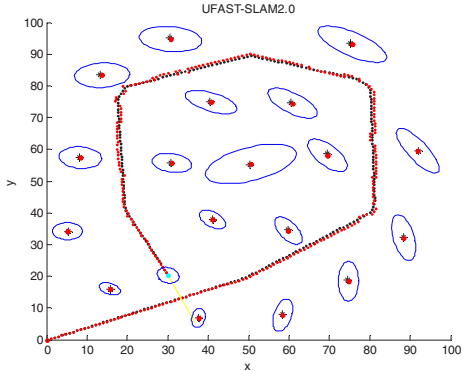

(b) Unscented FastSLAM

Fig. 2 The simulation results for $\sigma_r = 0.3m$, $\alpha_\varphi = 3^o$. The red and black dashed lines represent the reference trajectory and the estimate trajectory, respectively. The red dots and the black star represent the reference landmarks and the estimate landmarks, respectively.

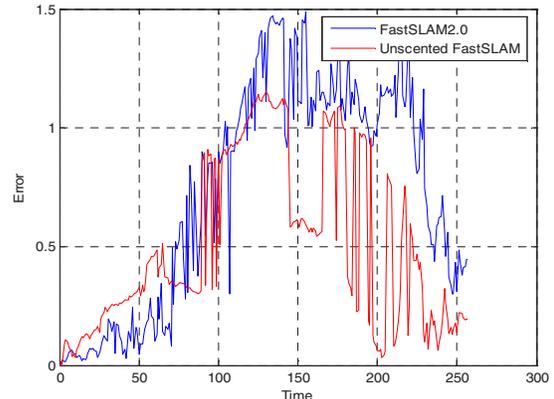

Fig. 3 The comparison of the pose error for $\sigma_r = 0.3m$, $\alpha_\varphi = 3^o$.

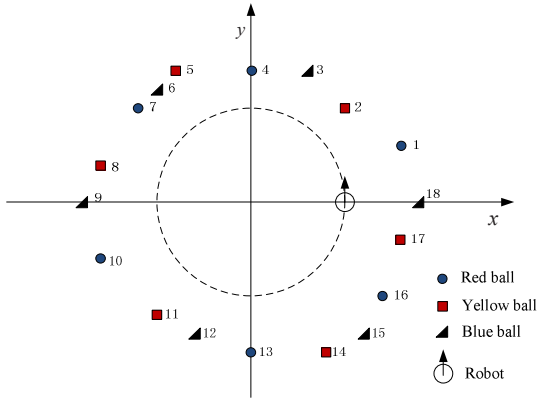

Fig. 4 Experiment scenario 1

*B. Experiments and discussion*

The experiments were conducted on the platform, a Pioneer 3-DX robot. A binocular camera system (MV-VD078SM/SC) was applied here to measure the range and the bearing of the landmark relative to the robot coordinate frame. The camera's resolution is 1024×768 pixels. The sampling period of camera is 2 second. The proposed technique has been evaluated comprehensively by two different types of the experiments.

Scenario 1: The robot moves along a circle trajectory in a small environment ($2m \times 2m$), Fig. 4. Eighteen balls with different color were used in this experiment as the landmarks. The robot moves with the speed $\pi/36$ *m/s* and the rotational velocity is $\pi/36$ *rad/s*.

Fig. 5 shows the experimental result of FastSLAM 2.0 and UFastSLAM. Fig. 6 shows the comparison of localization and mapping performance. According to Fig. 6, the robot pose error of UFastSLAM increased much more slowly than that of FastSLAM 2.0. The estimation of the landmark position using UFastSLAM is also smaller than FastSLAM2.0. Thus UFastSLAM can achieve more accurate performance than FastSLAM2.0.

Scenario 2: More generally, a relatively large environment ($26.7m \times 48m$) is selected to verify the superior performance of the UFastSLAM, Fig. 7. In Fig. 7, black blocks represent each room door and the dotted line represents the trajectory of robot where the robot moves along the long corridors. The experimental results of FastSLAM 2.0 and UFastSLAM are shown in Fig. 8. The experiment results show that the performance of vision based UFastSLAM is much better than using FastSLAM2.0 for long moving distance. Moreover, the estimated robot pose and landmark position using vision-based UFastSLAM coincide very well with the true trajectory.

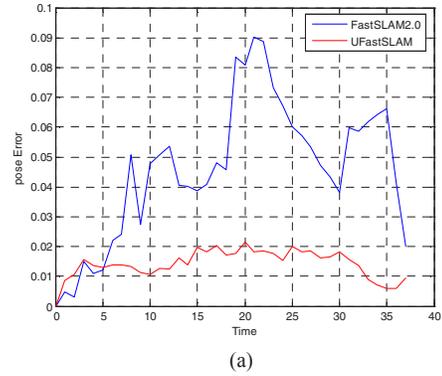

(a)

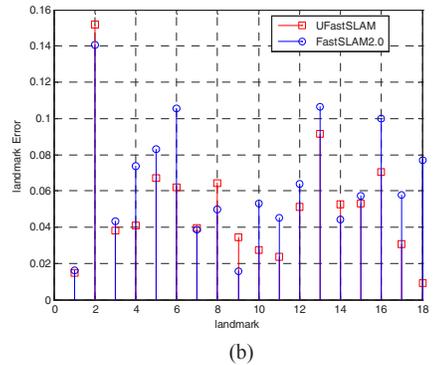

(b)

Fig. 6 The comparison of the estimation errors between two algorithms.

## V. CONCLUSIONS

This paper comprehensively investigates a vision-based Unscented FastSLAM for mobile robot. Simulations and experiments verify that the vision-based UFastSLAM is more robust and can improve the estimation accuracy comparing with FastSLAM2.0.

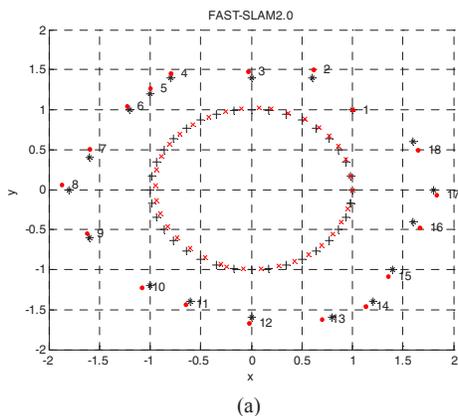

(a)

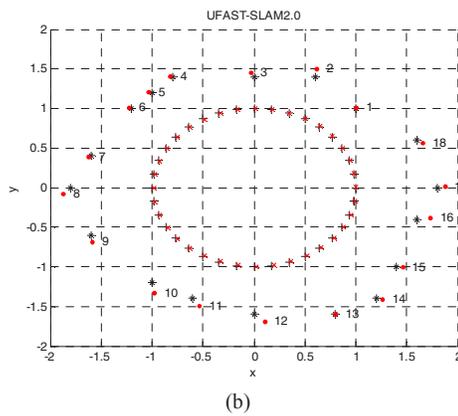

(b)

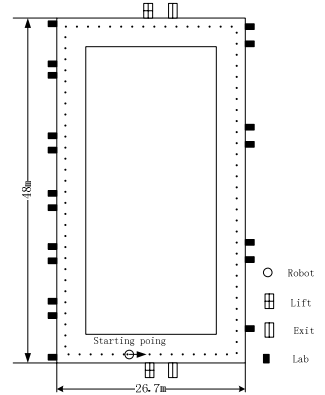

Fig. 7 Experiment scenario 2

Fig. 5 Experiment results- scenario 1. The red and black dashed lines represent the reference trajectory and the estimate trajectory, respectively. The red dots and the black stars represent the reference landmarks and the estimate landmarks, respectively.

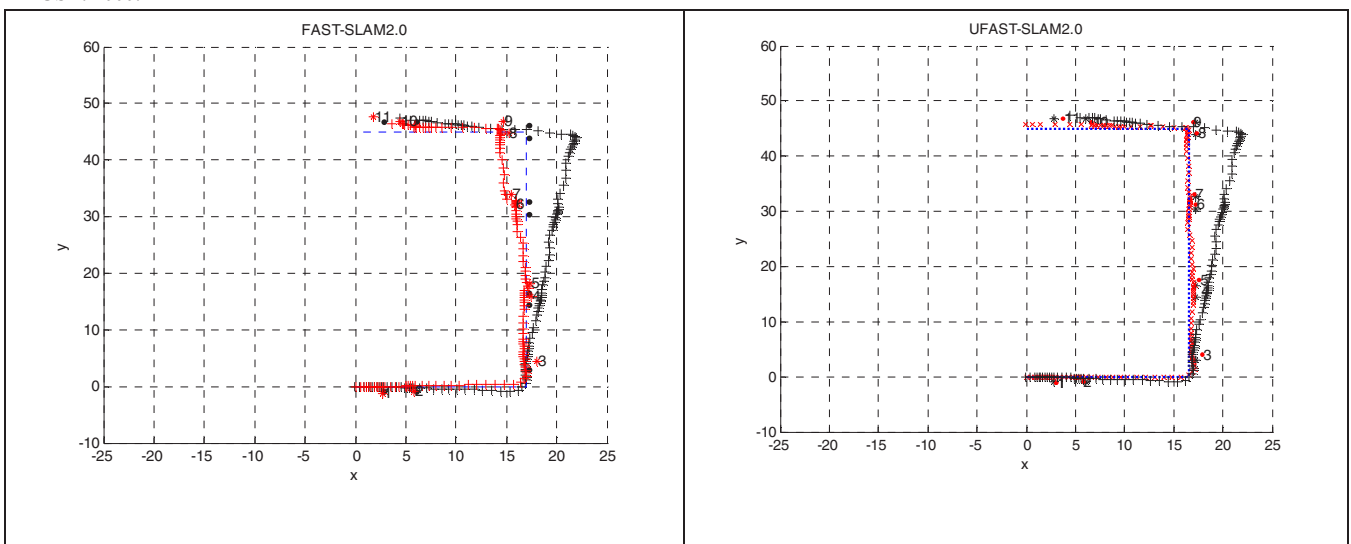

Fig. 8 Experimental result- scenario 2. The red plus line and the black star line represent the estimated trajectory the dead-reckoning, respectively. The blue dot line represents the true trajectory.